\documentclass[10pt,twocolumn,letterpaper]{article}

\usepackage{cvpr}
\usepackage{times}
\usepackage{epsfig}
\usepackage{graphicx}
\usepackage{amsmath}
\usepackage{amssymb}
\usepackage{fixltx2e}  
\usepackage[caption=false]{subfig}
\usepackage{caption}
\usepackage[absolute]{textpos}  
\usepackage{pbox}  

\usepackage[pagebackref=true,breaklinks=true,letterpaper=true,colorlinks,bookmarks=false]{hyperref}

\newcommand{\LSP}{LSP}
\newcommand{\LSPextended}{LSP-extended}
\newcommand{\MPII}{MPII-HumanPose}
\newcommand{\HumanEva}{HumanEva}
\newcommand{\HumanM}{Human3.6M}
\newcommand{\FP}{FashionPose}
\newcommand{\SMPLPose}{UP}

\newif\ifarxiv
\arxivtrue

{\cvprfinalcopy}  

\def\cvprPaperID{2644} 

\begin{document}

\title{
  Unite the People: Closing the Loop Between 3D and 2D Human Representations}
\author{Christoph Lassner\textsuperscript{1,2}\\
{\tt\small classner@tue.mpg.de}
\and
Javier Romero\textsuperscript{3,*}\\
{\tt\small javier.romero@bodylabs.com}
\and
Martin Kiefel\textsuperscript{2}\\
{\tt\small mkiefel@tue.mpg.de}
\and
Federica Bogo\textsuperscript{4,*}\\
{\tt\small febogo@microsoft.com}
\and
Michael J. Black\textsuperscript{2}\\
{\tt\small black@tue.mpg.de}
\and
Peter V. Gehler\textsuperscript{5,*}\\
{\tt\small pgehler@tue.mpg.de}
\vspace*{1.3cm}
}
\maketitle
\ifcvprfinal\ifarxiv\else\thispagestyle{empty}\pagestyle{empty}\fi\fi
{\ifcvprfinal}
	{\TPshowboxesfalse}
	\begin{textblock*}{120mm} (4cm, 7.3cm) {
      \noindent
      \begin{center}
        \textsuperscript{1}Bernstein Center for Computational Neuroscience, T\"ubingen, Germany\\
        \textsuperscript{2}MPI for Intelligent Systems, T\"ubingen, Germany\\
        \textsuperscript{3}Body Labs Inc., New York, United States\quad\quad
        \textsuperscript{4}Microsoft, Cambridge, UK\\
        \textsuperscript{5}University of W\"urzburg, Germany\\
      \end{center}
    }
  \end{textblock*}
\fi

\begin{abstract}
  \vspace*{-0.3cm}
  3D models provide a common ground for different representations of human
  bodies. In turn, robust 2D estimation has proven to be a powerful tool to
  obtain 3D fits ``in-the-wild''.
  However, depending on the level of detail, it can be hard to impossible
  to acquire labeled data for training 2D estimators on large scale.
  We propose a hybrid approach to this problem: with an extended version of the
  recently introduced SMPLify method, we obtain high quality 3D body model fits
  for multiple human pose datasets. Human annotators solely sort good and bad
  fits. This procedure leads to an initial dataset, \textit{UP-3D}, with rich
  annotations.
  With a comprehensive set of experiments, we show how this data can be used to
  train discriminative models that produce results with an unprecedented level of
  detail: our models predict 31 segments and 91 landmark locations on the body.
  Using the 91~landmark pose estimator, we present state-of-the art results for 3D human
  pose and shape estimation using an order of magnitude less
  training data and without assumptions about gender or pose in the fitting
  procedure. We show that \textit{UP-3D} can be enhanced with these improved
  fits to grow in quantity and quality,
  which makes the system deployable on large scale.
  The data, code and models are available for research purposes.
\end{abstract}
\renewcommand*{\thefootnote}{\fnsymbol{footnote}}
\footnotetext{\textsuperscript{*} This work was performed while J. Romero and F. Bogo were with the MPI-IS\textsuperscript{2}; P. V. Gehler with the BCCN\textsuperscript{1} and MPI-IS\textsuperscript{2}.}
\renewcommand*{\thefootnote}{\arabic{footnote}}
\setcounter{footnote}{0}

\begin{figure}
  \vspace*{0.2cm}
  \hfill
  \includegraphics[width=0.48\textwidth]{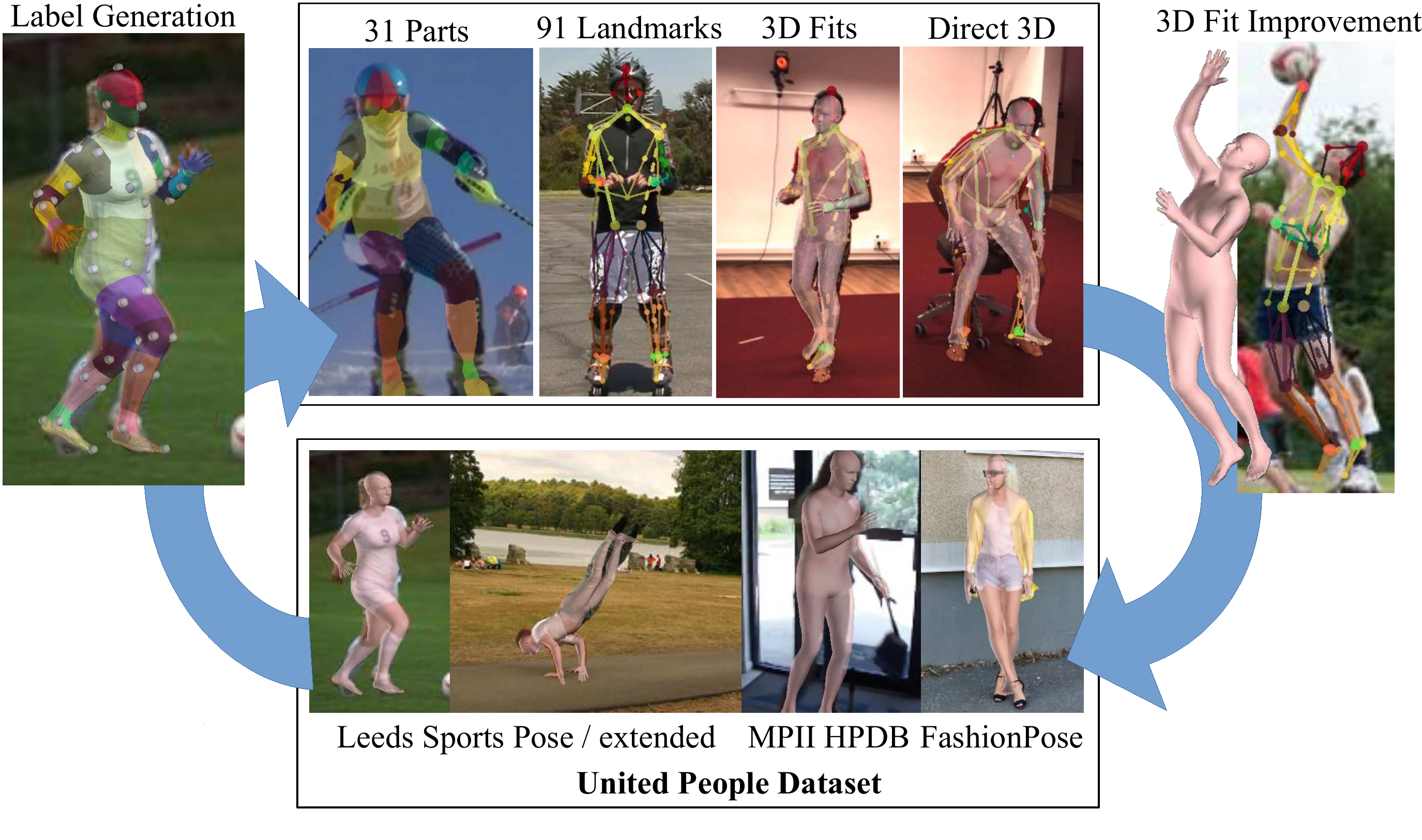}
  \hfill
  \vspace*{-0.8cm}
  \caption{\textbf{Lower row:} validated 3D body model fits on 
    various datasets form our initial dataset, \textit{UP-3D}, and
    provide labels for multiple tasks. \textbf{Top row:} we perform experiments
    on semantic body part segmentation, pose estimation and 3D fitting.
    Improved 3D fits can extend the initial dataset. 
    \label{fig:nugget}}
  \vspace*{-0.2cm}
\end{figure}

\section{Introduction}

Teaching computers to recognize and understand humans in images and videos is a
fundamental task of computer vision.
Different applications require
different trade-offs between
fidelity of the representation and
inference complexity.
This led to a wide range of parameterizations for
human bodies
and corresponding prediction methods
ranging from bounding boxes
to detailed 3D models.

Learning-based algorithms, especially convolutional neural networks (CNNs), are
the leading methods to cope with the complexity of human appearance.
Their representational power has led to increasingly robust algorithms for bounding box
detection~\cite{dollar2012pedestrian}, keypoint
detection~\cite{deepercut,Newell16,ConvolutionalPoseMachines:2016} and body part
segmentation~\cite{Chen2016,Hariharan2015,Xia2016}. However, they are usually applied in
isolation on separate datasets and independent from the goal of precise 3D body estimation. In this
paper we aim to overcome this separation and ``unite the people'' of different
datasets 
and for multiple tasks.
With this strategy, we attack the main problem of learning-based approaches for
complex body representations:
the lack of data. While it is feasible to annotate a small number of keypoints
in images (\eg, 14 in the case of the {\MPII} dataset~\cite{mpii}), scaling to
larger numbers quickly becomes impractical and prone to annotation inconsistency.
The same is true for semantic segmentation annotations: most datasets provide labels for
only a few body parts.


In this paper, we aim to develop a self-improving, scalable method that obtains
high-quality 3D body model fits for 2D images (see Fig.~\ref{fig:nugget} for an
illustration).
To form an initial dataset of 3D body fits, we use an improved
version of the recently developed SMPLify method~\cite{paper2} that elevates 2D
keypoints to a full body model of pose and shape. A more robust initialization
and an additional fitting objective allow us to apply it on the ground truth
keypoints of the standard human pose datasets; human annotators solely sort good
and bad fits.

This semi-automatic scheme has several advantages.
The required annotation time is greatly reduced
(Sec.~\ref{sec:exploring}). By projecting surfaces (Sec.~\ref{sec:segmentation})
or keypoints (Sec.~\ref{sec:pose}) from the fits to the original images, we
obtain 
consistent labels while retaining generalization
performance. The rich representation and the flexible fitting process make it
easy to integrate datasets with different label sets, \eg, a different set of
keypoint locations.

Predictions from our 91 keypoint model improve the 3D model fitting method
that generated the annotations for training the keypoint model in the first
place. We report state-of-the art results on the {\HumanEva}
and Human3.6M datasets (Sec.~\ref{sec:3d_pose_estimation}). Further, using the
3D body fits, we develop a random forest method for 3D pose estimation that
runs orders of magnitudes faster than SMPLify (Sec.~\ref{sec:dp}).

The improved predictions from the 91~landmark model increase the ratio of
high quality 3D fits on the LSP dataset by 9.3\% when compared to the fits
using 14~keypoint \emph{ground truth} locations (Sec.~\ref{sec:closingtheloop}).
This ability for self-improvement together with the possibility to easily
integrate new data into the pool
make the presented system deployable on large scale. Data, code and models
are available for research purposes on the project homepage at
\url{http://up.is.tuebingen.mpg.de/}.

\section{Related Work}

Acquiring human pose annotations in 3D is a long-standing problem with several
attempts from the computer vision as well as the 3D human pose community.

The classical 2D representation of humans are 2D
keypoints~\cite{mpii,Charles16,LSP:2010,lspext,modec13,sapp2011parsing}. While 2D keypoint
prediction has seen considerable progress in the last years and could be
considered close to being solved~\cite{deepercut,Newell16,ConvolutionalPoseMachines:2016}, 3D pose estimation from single images remains a
challenge~\cite{paper2,Ramakrishna:2012,Zhou:2010}.

Bourdev and Malik~\cite{PoseletsICCV09} enhanced the H3D dataset from 20
keypoint annotations for 1,240 people in 2D with relative 3D information as
well as 11 annotated body part segments.

In contrast, the {\HumanEva}~\cite{Sigal:2010} and
{\HumanM}~\cite{Human36m:2014} datasets provide very accurate 3D labels: they
are both recorded in motion capture environments. Both datasets have high fidelity
but contain only a very limited level of diversity in background and
person appearance. We evaluate the 3D human pose estimation performance on both.
Recent approaches target 3D pose ground truth from natural scenes, but
either rely on vision systems prone to failure~\cite{elhayek2015efficient}
or inertial suits that modify the appearance of the body and are prone to motion
drift~\cite{Zhou:2010}.

Body representations beyond 3D skeletons have a long history in the
computer vision community~\cite{HOGG19835,Marr269,Nevatia:1973,pons-moll:mrf}. More recently,
these representations have taken new popularity in approaches that fit
detailed surfaces of a body model to
images~\cite{paper2,Guan:2009,Hasler:2010,li20143d,Zhou:2010}. These representations
are more tightly connected to the physical reality of the human body and the
image formation process.

One of the classic problems related to representations of the extent of the body
is body part segmentation. Fine-grained part segmentation has been added to the
public parts of the VOC dataset~\cite{Everingham:2010} by Chen et
al.~\cite{chen_cvpr14}. Annotations for 24 human body parts and also part
segments for all VOC object classes, where applicable, are available. Even though hard to
compare, we provide results on the dataset. The Freiburg Sitting People
dataset~\cite{Oliveira:ICRA:2016} consists of 200 images with 14 part
segmentation and is tailored towards sitting poses. The ideas by Shotton et
al.~\cite{kinect} for 2.5D data inspired our body part representation.
Relatively simple methods have proven to achieve
good performance in segmentation tasks with ``easy'' backgrounds like
Human80k, a subset of Human3.6M~\cite{ionescu2014iterated}.

Following previous work on cardboard
people~\cite{Black:ICAFGR:1996} and contour people~\cite{Freifeld:CVPR:10}, an
attempt to work towards an intermediate-level person representation is the JHMDB
dataset and the related labeling tool~\cite{Jhuang:ICCV:2013}. It relies on
`puppets' to ease the annotation task, while providing a higher level of detail
than solely joint locations.


The attempt to unify representations for human bodies has been made
mainly in the context of human kinematics~\cite{azad:IROS:2007,
  Mandery2015_KITMotionDB}. In their work, a rich representation for 3D motion capture
marker sets is used to transfer captures to different targets. The setup of
markers to capture not only human motion but also shape has been explored by
Loper et al.~\cite{Loper:SIGASIA:2014} for motion capture scenarios. While they optimized the placement of
markers for a 12 camera setup, we must ensure that the markers
disambiguate pose and shape from a single view. Hence, we use a
denser set of markers.

\section{Building the Initial Dataset}

Our motivation to use a common 3D representation is to (1) map many
possible representations from a variety of datasets to it, and (2) generate
detailed and consistent labels for supervised model training from it.

We argue that the use of a full human body model with a prior on shape
and pose is necessary: without the visualization possibilities and regularization, it may be impossible to create sufficiently accurate annotations
for small body parts. 
However, so far, no dataset is available that provides human body model fits on a
large variety of images.

To fill this gap, we build on a set of human pose datasets with annotated keypoints. 
SMPLify~\cite{paper2} presented promising results for automatically translating these into 3D
body model fits. 
This helps us to keep the human involvement to
a minimum. With strongly increasing working times 
and levels of label noise for increasingly complex tasks,
this may be a critical decision to
create a large dataset of 3D body models.

\subsection{Improving Body Shape Estimation\label{sec:segmentation_term}}

In~\cite{paper2}, the authors fit the pose and shape parameters of the
SMPL~\cite{SMPL:2015} body model to 2D keypoints by minimizing an objective
function composed of a data term and several penalty terms that represent priors
over pose and shape. However, the connection length between two keypoints is the
only indicator that can be used to estimate body shape. Our aim is to match the
shape of the body model as accurately as possible to the images, hence we must
incorporate a shape objective in the fitting.

The best evidence for the extent of a 3D body projected on a 2D image is encoded
by its silhouette. We define the silhouette to be the set of all pixels
belonging to a body's projection. Hence, we add a term to the original SMPLify
objective to prefer solutions for which the image silhouette, $S$, and the model
silhouette, $\hat{S}$, match.

Let $M(\vec{\theta}, \vec{\beta}, \vec{\gamma})$ be a 3D mesh generated by a
SMPL body model with pose, $\vec{\theta}$, shape, $\vec{\beta}$, and global
translation, $\vec{\gamma}$. Let $\Pi(\cdot, K)$ be a function that takes a 3D
mesh and projects it into the image plane given camera parameters $K$, such that
$\hat{S}(\vec{\theta}, \vec{\beta}, \vec{\gamma})=\Pi(M(\vec{\theta},
\vec{\beta}, \vec{\gamma}))$ represents the silhouette pixels of the model in
the image.
We compute the bi-directional distance between $S$ and $\hat{S}(\cdot)$
\begin{eqnarray}
E_{S}(\vec{\theta},\vec{\beta},\vec{\gamma};S,K) & = & \sum_{\vec{x}\in\hat{S}(\vec{\theta},\vec{\beta},\vec{\gamma})}\mathrm{dist}{(\vec{x},S)}^{2}\nonumber \\
 &  & +\sum_{\vec{x}\in S}\mathrm{dist}(\vec{x},\hat{S}(\vec{\theta},\vec{\beta},\vec{\gamma})),\label{eq:smpl}
\end{eqnarray}
where $\mathrm{dist}(\vec{x},S)$ denotes the absolute distance from a point
$\vec{x}$ to the closest point belonging to the silhouette
$S$. 

The first term in Eq.~\eqref{eq:smpl} computes the distance from points of the
projected model to a given silhouette, while the second term computes the
distance from points in the silhouette to the model. We find that the second
term is noisier and use the plain L1 distance to measure its contribution to the
energy function while we use the squared L2 distance to measure the
contribution of the first. We optimize the overall objective including this
additional term using OpenDR~\cite{Loper:2014}, just as in~\cite{paper2}.

\begin{table}
  \begin{centering}
    \begin{tabular}{|c|c|c|c|}
      \hline
      {\scriptsize{}Dataset} & {\scriptsize{}Foreground} & {\scriptsize{}6 Body Parts} & {\scriptsize{}AMT hours logged} \\
      \hline
      {\scriptsize{}LSP~\cite{LSP:2010}} & {\scriptsize{}\pbox{20cm}{1000 train,\\1000 test{\scriptsize{}}}} & {\scriptsize{}\pbox{20cm}{1000 train,\\1000 test{\scriptsize{}}}} & {\scriptsize{}361h foreground},\\
      \cline{1-3}
      {\scriptsize{}LSP-extended~\cite{lspext}} & {\scriptsize{}10000 train} & {\scriptsize{}0} & {\scriptsize{}131h parts} \\
      \hline
      {\scriptsize{}MPII-HPDB~\cite{mpii}} & {\scriptsize{}\pbox{20cm}{13030 train,\\2622 test{\scriptsize{}}}} & {\scriptsize{}0} & {\scriptsize{}729h}\\
      \hline
    \end{tabular}
    \par\end{centering}
  \vspace*{-0.2cm}
  \caption{Logged AMT labelling times. The average foreground labeling task was solved in 108s on the LSP and
    168s on the MPII datasets respectively. Annotating the segmentation for six body parts took on average more than twice as long as annotating foreground segmentation:~236s.\label{tab:AMT_times}}
\end{table}

\begin{figure}
\hfill
\includegraphics[height=2.2cm]{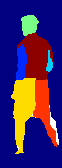}\includegraphics[height=2.2cm]{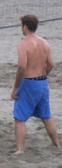}
\hfill
\includegraphics[height=2.2cm]{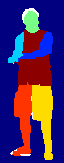}\includegraphics[height=2.2cm]{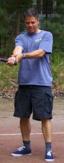}
\hfill
\includegraphics[height=2.2cm]{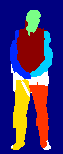}\includegraphics[height=2.2cm]{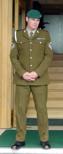}
\hfill
\includegraphics[height=2.2cm]{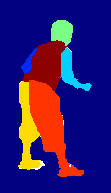}\includegraphics[height=2.2cm]{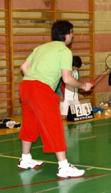}
\hfill
\vspace*{-0.2cm}
\caption{Examples for six part segmentation ground truth. White areas
mark inconsistencies with the foreground segmentation and are ignored.}
\label{fig:seg_annotations}
\vspace*{-0.2cm}
\end{figure}

Whereas it would be possible to use an
automatic segmentation method to provide foreground silhouettes, we decided to involve human annotators for reliability. We also asked for six body part segmentation that we will use in Sec.~\ref{sec:experiments} for evaluation. 
We built an interactive annotation tool on top of the
\textit{Opensurfaces} package~\cite{Opensurfaces:2013} to work with Amazon
Mechanical Turk (AMT)\@. To obtain image-consistent silhouette borders, we use
the interactive Grabcut algorithm~\cite{rother04grabcut}. Workers spent more
than 1,200 hours on creating the labels for the {\LSP}~\cite{LSP:2010,lspext} datasets
as well as the single-person part of the {\MPII}~\cite{mpii} dataset (see
Tab.~\ref{tab:AMT_times}). There is an increase in average annotation
time of more than a factor of two comparing annotation for foreground labels and six body
part labels. This provides a hint on how long annotation for a 31 body part
representation could take. Examples for six part segmentation labels are
provided in Fig.~\ref{fig:seg_annotations}.

\subsection{Handling Noisy Ground Truth Keypoints}


The SMPLify method is
especially vulnerable to missing annotations of the four torso joints: it
uses their locations for an initial depth guess, and convergence deteriorates
if this guess is of poor quality.

Finding a good depth initialization is particularly hard due to the
foreshortening effect of the perspective projection. However, since we know that
only a shortening but no lengthening effect can occur, we can find a more reliable person size estimate $\hat{\theta}$
for a skeleton model with $k$ connections:
\begin{equation}
  \hat\theta=\mathbf{x}_i\cdot \underset{y}{\arg\max} f_i(y),\;\;
  i=\underset{j=1,\ldots,k}{\arg\max}\,\mathbf{x}_j,
\end{equation}
where $f_i$ is the distribution over ratios of person size to the length of
connection $\textbf{x}_i$. Since this is a skewed distribution, we use a
corrected mean to find the solution of the $\arg\max$ function and obtain a
person size estimate. This turns out to be a simple, yet robust estimator.

\subsection{Exploring the Data\label{sec:exploring}}

\begin{figure*}
  \vspace*{-0.4cm}
  \subfloat[\label{fig:human_labels_keypoints}]{
    \includegraphics[height=4cm]{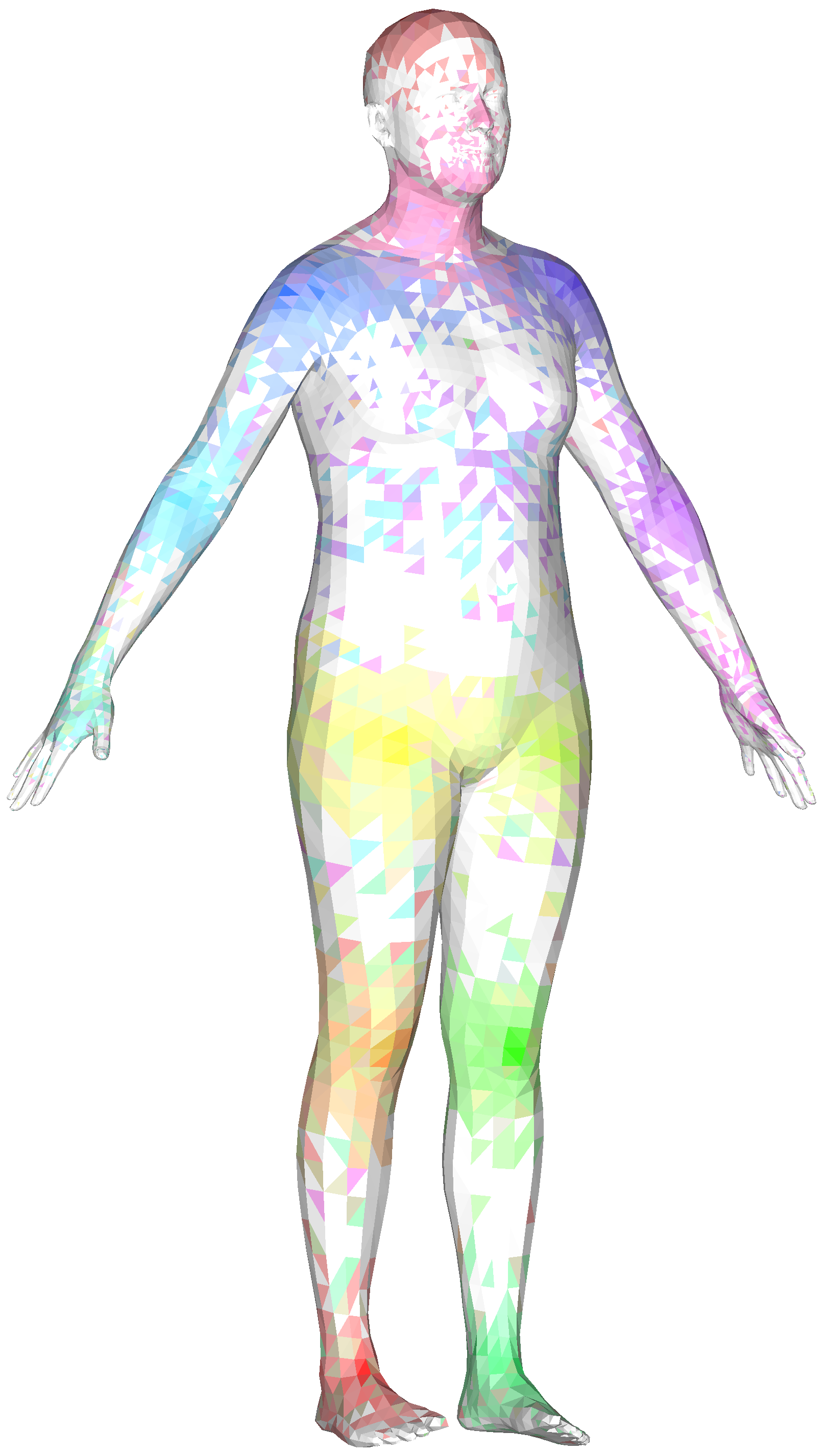}
    \includegraphics[height=4cm]{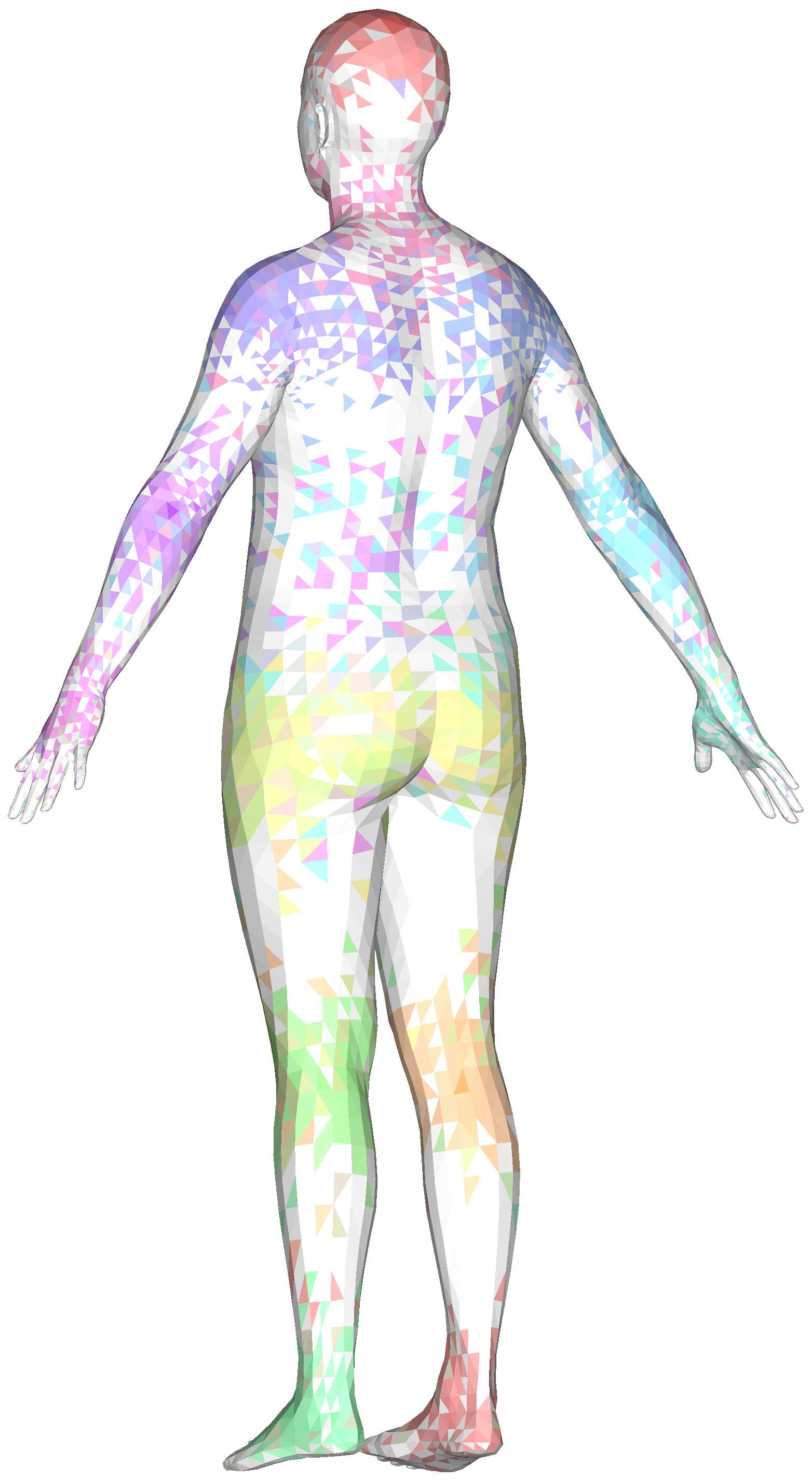}
  }
  \hfill
  \subfloat[\label{fig:human_labels_seg}]{
    \includegraphics[height=4cm]{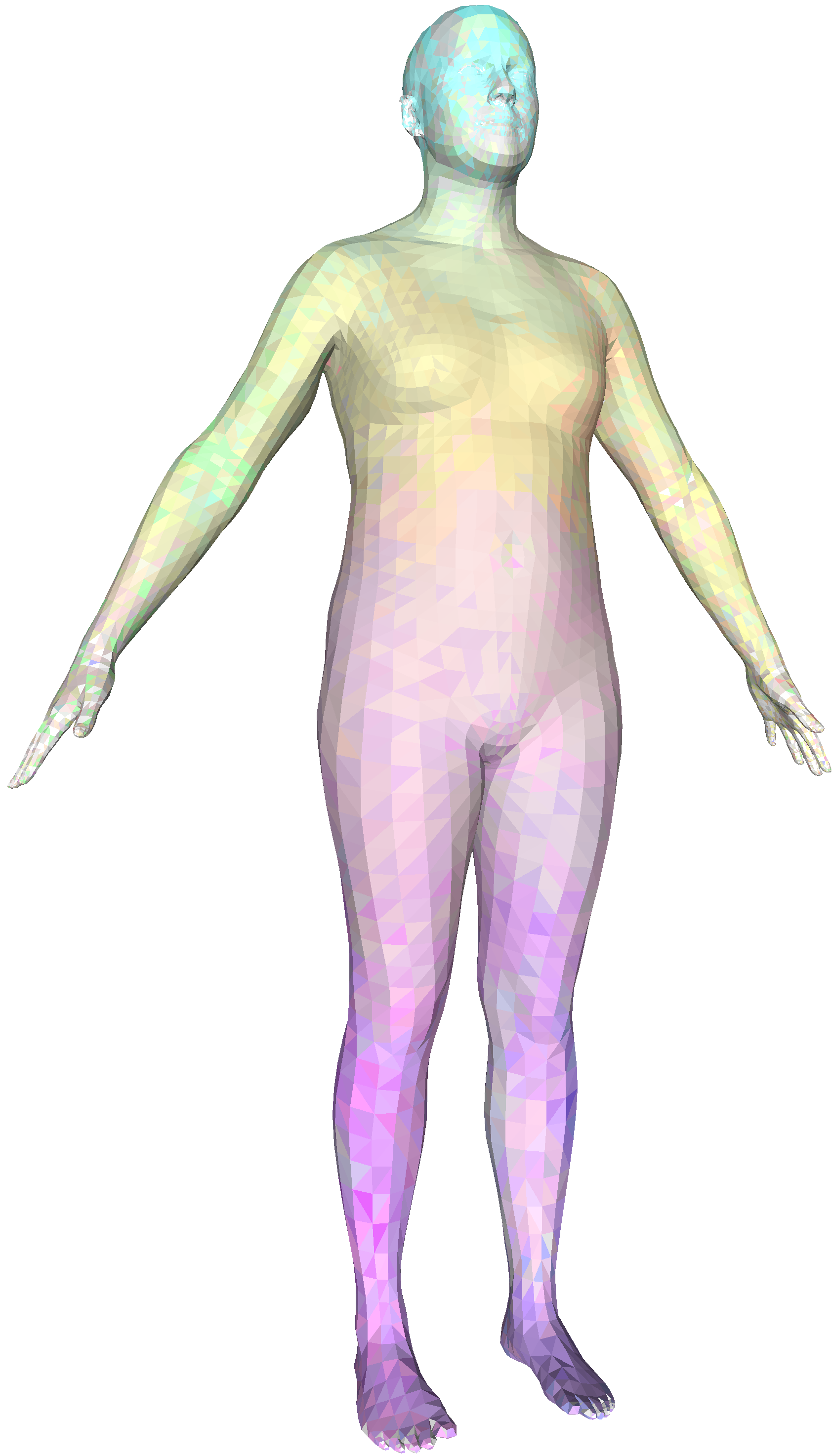}
    \includegraphics[height=4cm]{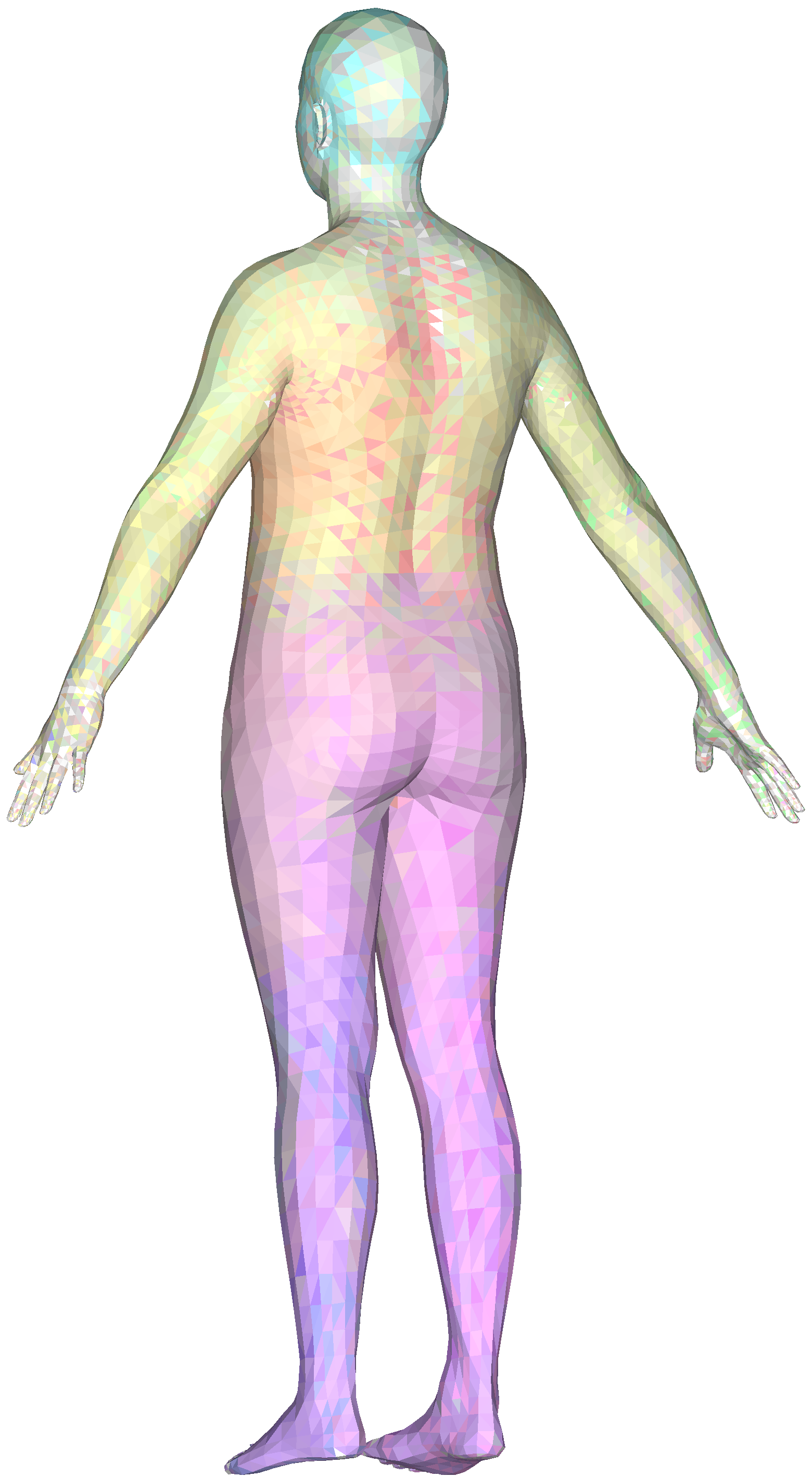}
  }
  \hfill
  \subfloat[\label{fig:landmarks}]{
    \includegraphics[height=4cm]{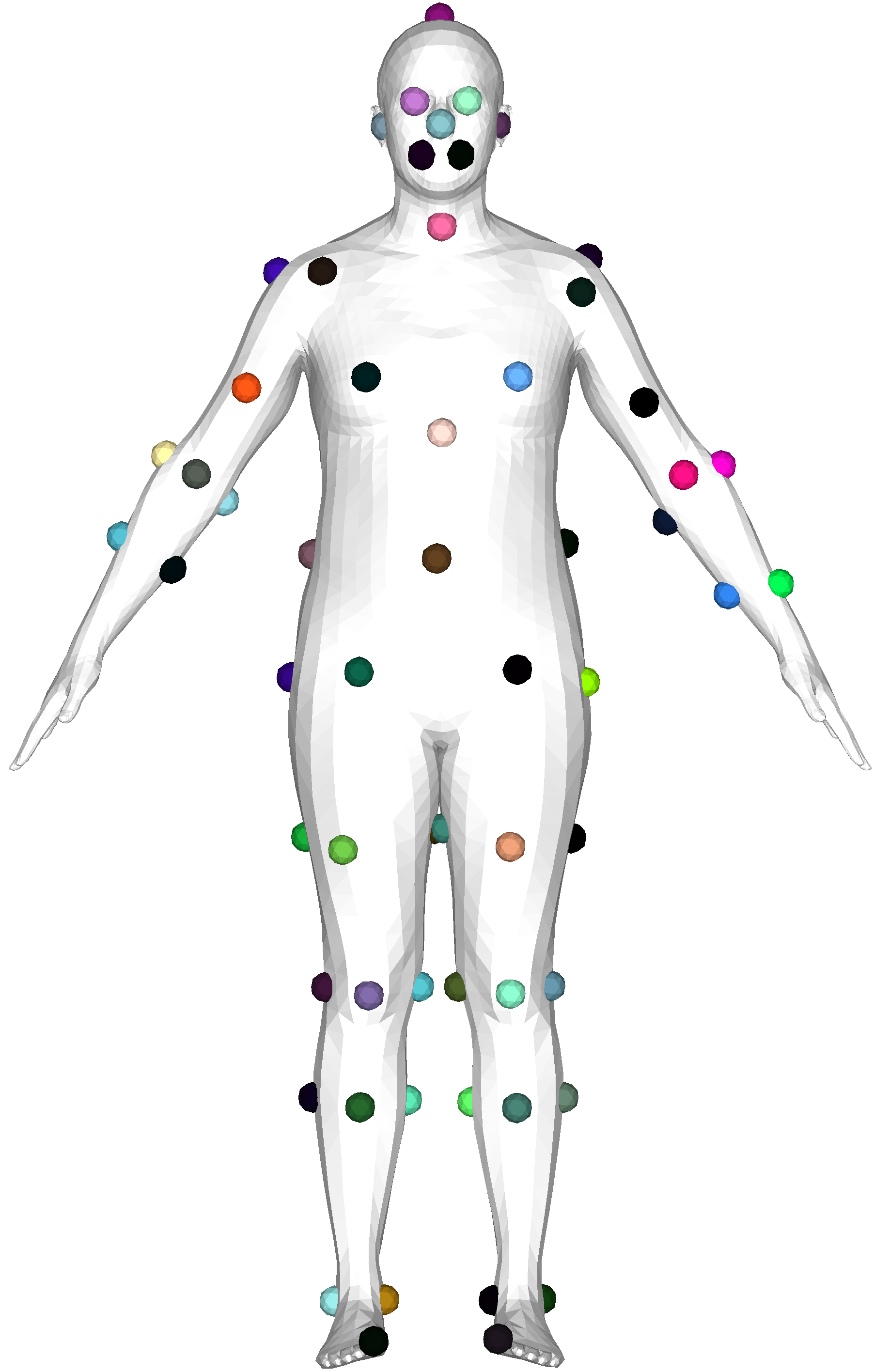}\includegraphics[height=4cm]{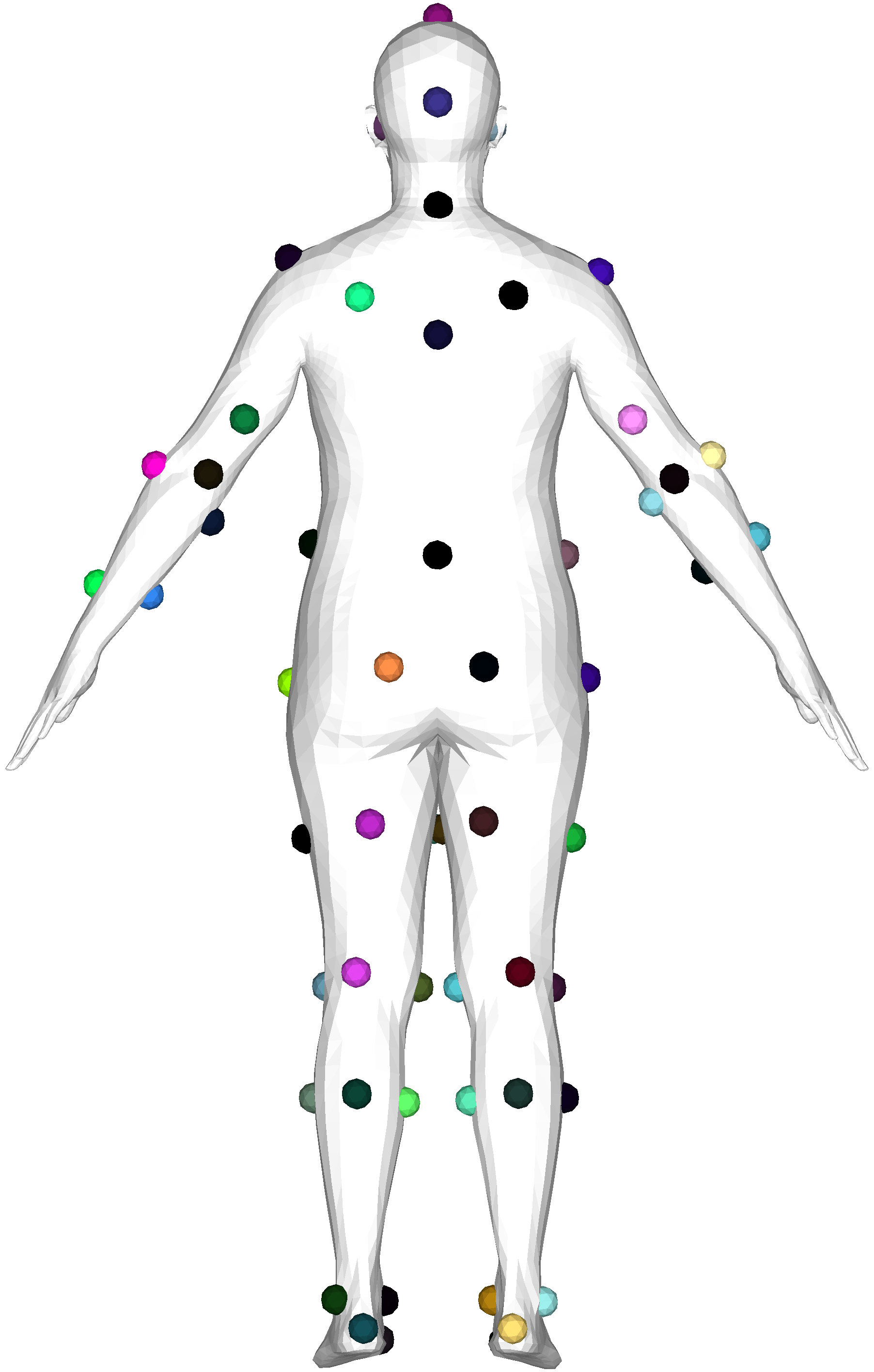}}
  \hfill
  \subfloat[\label{fig:segmentation_gt}]{\includegraphics[height=4cm]{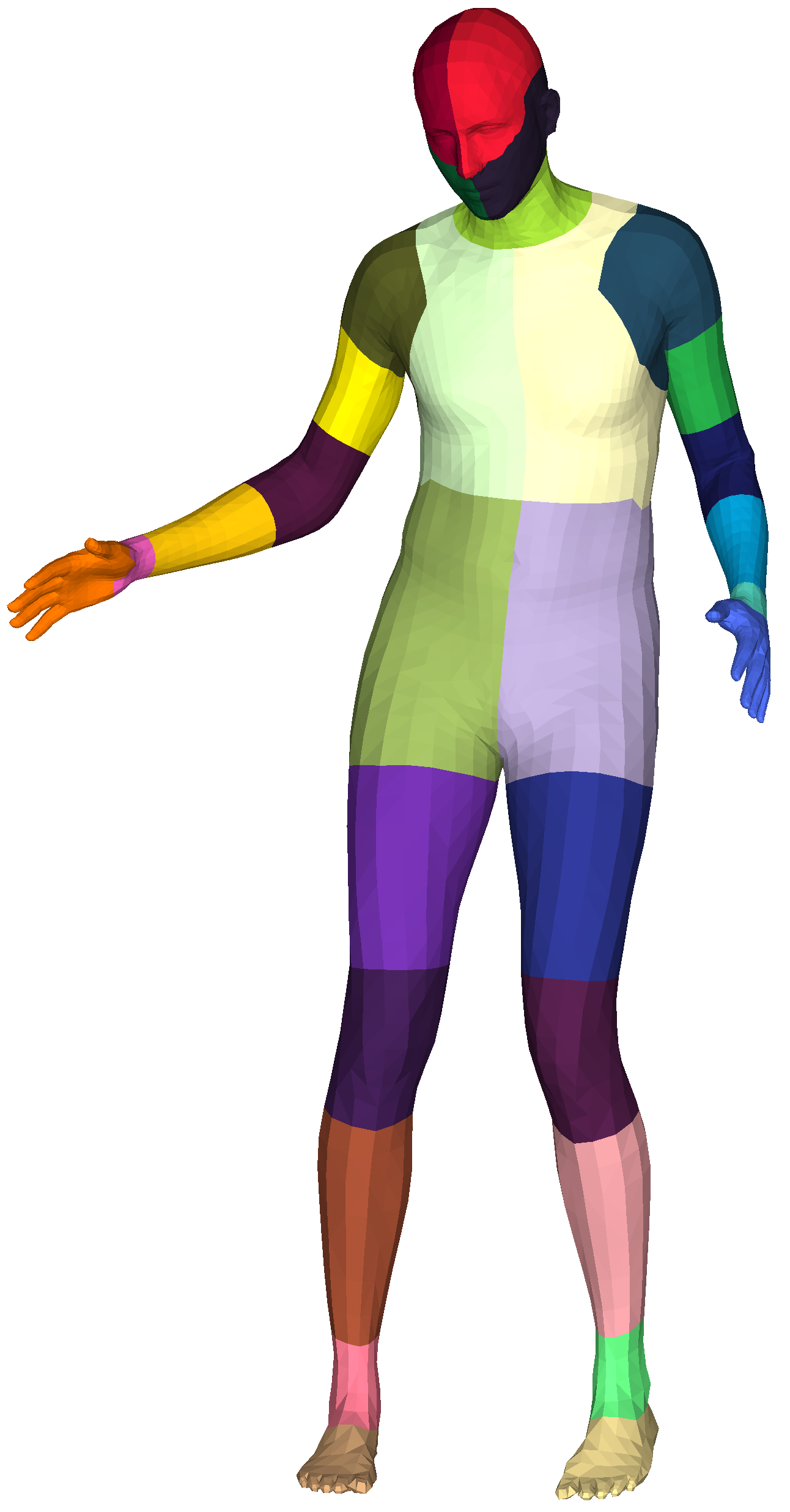}}
  \hfill
  \vspace*{-0.2cm}
  \caption{Density of human annotations on high quality body model fits for
    \textbf{(a)}~keypoints and \textbf{(b)}~six part segmentation in front and
    back views. Areas of the bodies are colored with (1) hue according to part
    label, and (2) saturation according to frequency of the label. Keypoints on
    completely `wrong' bodyparts are due to self-occlusion. The high
    concentration of `head' labels in the nose region originates from the {\FP}
    dataset, where the `head' keypoint is placed on the nose. The segmentation
    data originates solely from the six part segmentation labels on the {\LSP}
    dataset. (Must be viewed in color.) \textbf{(c)} Placement of the 91
    landmarks (left: front, right: back). \textbf{(d)} Segmentation for generating the 31 part labels.\label{fig:human_labels}}
\end{figure*}

With the foreground segmentation data and the
adjustments described in the preceding sections, we fit the SMPL model to a
total of 27,652 images of the {\LSP}, {\LSPextended}, and {\MPII} datasets. We
use only people marked with the `single person' flag in {\MPII} to avoid
instance segmentation problems. We honor the train/test splits of the datasets
and keep images from their test sets in our new, joined test set.

In the next step, human annotators\footnote{For this task, we did not rely on AMT workers, but only on few experts in close collaboration to maintain consistency.} selected the fits
where rotation and location of body parts largely match the image evidence. For
this task, we provide the original image, as well as four perspectives of
renderings of the body. Optionally, annotators can overlay rendering and image.
These visualizations help to identify fitting errors quickly and reduce the
labeling time to $\sim$12s per image. The process uncovered many
erroneously labeled keypoints, where mistakes in the 3D fit were clear to spot,
but not obvious in the 2D representation. We excluded head and foot rotation
as criteria for the sorting process. There is usually not sufficient information in
the original 14 keypoints to estimate them correctly. The resulting ratios of accepted fits can be found in Tab.~\ref{tab:usable_fits}.

\begin{table}
  \begin{centering}
    \resizebox{0.48\textwidth}{!}{
    \begin{tabular}{|c|c|c|c|}
      \hline
      \scriptsize{}LSP~\cite{LSP:2010}& \scriptsize{}LSP extended~\cite{lspext} &\scriptsize{} MPII-HP~\cite{mpii} & \scriptsize{}FashionPose~\cite{Fashionpose:2014}\\
      \hline
      \hline
      \scriptsize{}45\% &\scriptsize{} 12\% & \scriptsize{}25\% &\scriptsize{} 23\%\\
      \hline
    \end{tabular}
    \par
  }\end{centering}
  \vspace*{-0.2cm}
  \caption{Percentages of accepted fits per dataset. The addition of the FashionPose dataset
    is discussed in Sec.~\ref{sec:integrating_new_data}.\label{tab:usable_fits}}
  \vspace*{-0.2cm}
\end{table}

Even with the proposed, more robust initialization term, the ratio of accepted fits on the
{\LSPextended} dataset remains the lowest. It has the highest number of missing
keypoints of the four datasets, and at the same time the most 
extreme viewpoints and poses. On the other hand, the rather high
ratio of usable fits on the {\LSP} dataset can be explained with the clean and
complete annotations.

The validated fits form our initial dataset with 5,569 training images (of which
we use a held-out validation set of 1,112 images in our experiments) and 1,208
test images. We denote this dataset as {\SMPLPose}I-3D ({\bf U}nited{\bf P}eople
in 3D with an added `I' for ``Initial''). To be able to clearly reference the
different label types in the following sections, we add an `h' to the dataset
name when referring to labels from human annotators.

\paragraph{Consistency of Human Labels}
The set of curated 3D fits allows us to assess the distribution of the
human-provided labels by projecting them to the {\SMPLPose}I-3D bodies. We did this for
both, keypoints and body part segments. Visualizations can be found in
Fig.~\ref{fig:human_labels}.

While keypoint locations in Fig.~\ref{fig:human_labels_keypoints} in completely
non-matching areas of the body can be explained by self-occlusion, 
there is a high variance in keypoint locations around joints. It must be taken
into account that the keypoints 
are projected to the body surface, and depending on person shape and
body part orientation some variation can be expected. Nevertheless, even for
this reduced set of images with very good 3D fits,
high variance areas, \eg, around the hip joints, indicate labeling noise.

The visualization in Fig.~\ref{fig:human_labels_seg} shows the density of part
types for six part segmentation with the segments head, torso, left and right arms and
left and right legs. While the head and lower parts of the extremities resemble
distinct colors, the areas converging to brown represent a mixture of part
annotations. The brown tone on the torso is a clear indicator for the frequent
occlusion by the arms. The area around the hips is showing a smooth transition
from torso to leg color, hinting again at varying annotation styles.

\section{Label Generation and Learning\label{sec:experiments}}

In a comprehensive series of experiments, we analyze the quality
of labels generated from {\SMPLPose}I-3D. We focus on labels for well-established
tasks, but highlight that the generation possibilities are not limited to them:
all types of data that can be extracted from the body model can be used as
labels for supervised training. In our experiments, we move from surface
(segmentation) prediction over 2D- to 3D-pose and shape estimation to a 
method for predicting 3D body pose and shape directly from 2D landmark positions.

\subsection{Semantic Body Part Segmentation\label{sec:segmentation}}

We segment the SMPL mesh into 31 regions, following the
segmentation into semantic parts introduced in~\cite{kinect} (for a
visualization, see Fig.~\ref{fig:segmentation_gt}). We note that the Kinect
tracker works on 2.5D data while our detectors only receive 2D data as
input. We deliberately did not make any of our methods for data collection
or prediction dependent on 2.5D data to retain generality. This way, we can
use it on outdoor images and regular 2D photo datasets. 
The {\bf S}egmentation dataset {\SMPLPose}I-S31 is obtained by projecting the
segmented 3D mesh posed on the 6,777 images of {\SMPLPose}I-3D.

\begin{figure*}
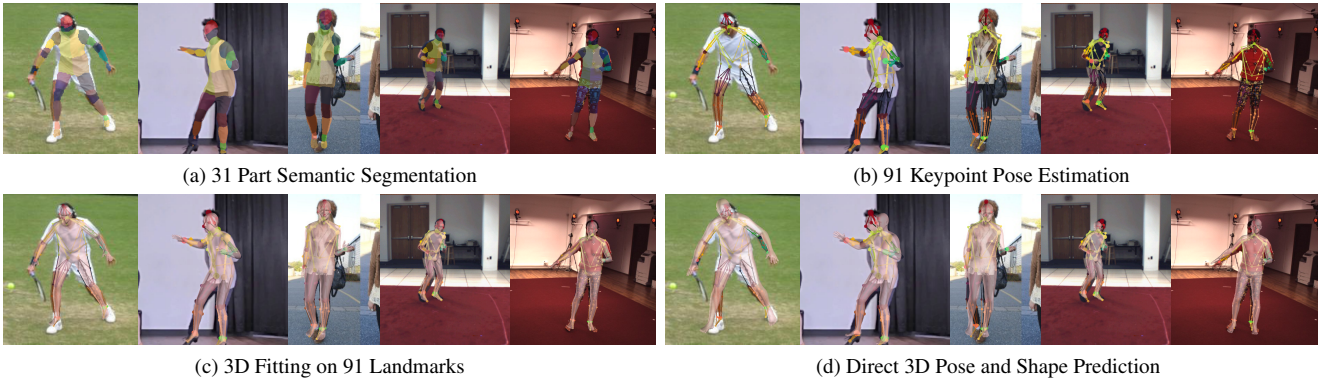

\subfloat[31 Part Semantic Segmentation\label{fig:results_seg}]{%
\includegraphics[height=2cm]{{{figures/results/part_segmentation/im1131.jpg_psegmentation.npz_vis}}}%
\includegraphics[height=2cm]{{{figures/results/part_segmentation/10009_image.png_psegmentation.npz_vis}}}%
\includegraphics[height=2cm]{{{figures/results/part_segmentation/15018_image.png_psegmentation.npz_vis}}}%
\includegraphics[height=2cm]{{{figures/results/part_segmentation/frame0157.png_psegmentation.npz_vis_cropped}}}%
\includegraphics[height=2cm]{{{figures/results/part_segmentation/frame2115.png_psegmentation.npz_vis}}}%
}%
\hfill\hspace*{0.12cm}%
\subfloat[91 Keypoint Pose Estimation\label{fig:results_pose}]{%
\includegraphics[height=2cm]{{{figures/results/pose/im1131.jpg_pose}}}%
\includegraphics[height=2cm]{{{figures/results/pose/10009_image.png_pose}}}%
\includegraphics[height=2cm]{{{figures/results/pose/15018_image.png_pose}}}%
\includegraphics[height=2cm]{{{figures/results/pose/frame0157.png_pose_cropped}}}%
\includegraphics[height=2cm]{{{figures/results/pose/frame2115.png_pose}}}%
}
\\[-2ex]%
\subfloat[3D Fitting on 91 Landmarks\label{fig:results_3d}]{%
\includegraphics[height=2cm]{{{figures/results/3D/im1131.jpg_body.pkl.overlay}}}%
\includegraphics[height=2cm]{{{figures/results/3D/10009_image.png_body.pkl.overlay}}}%
\includegraphics[height=2cm]{{{figures/results/3D/15018_image.png_body.pkl.overlay}}}%
\includegraphics[height=2cm]{{{figures/results/3D/frame0157.png_body.pkl.overlay_cropped}}}%
\includegraphics[height=2cm]{{{figures/results/3D/frame2115.png_body.pkl.overlay}}}%
}%
\hfill%
\subfloat[Direct 3D Pose and Shape Prediction\label{fig:results_d3d}]{
\includegraphics[height=2cm]{{{figures/results/direct/im1131.jpg_body_directseparate.pkl.overlay}}}%
\includegraphics[height=2cm]{{{figures/results/direct/10009_image.png_body_directseparate.pkl.overlay}}}%
\includegraphics[height=2cm]{{{figures/results/direct/15018_image.png_body_directseparate.pkl.overlay}}}%
\includegraphics[height=2cm]{{{figures/results/direct/frame0157.png_body_directseparate.pkl.overlay_cropped}}}%
\includegraphics[height=2cm]{{{figures/results/direct/frame2115.png_body_directseparate.pkl.overlay}}}%
}%
\\%
  \vspace*{-0.2cm}
  \caption{Results from various methods trained on labels generated from the
    {\SMPLPose}-3D
    dataset.}
  \label{fig:results}
  \vspace*{-0cm}
\end{figure*}

Following~\cite{Chen2016}, we optimize a multiscale ResNet101 on a pixel-wise
cross entropy loss. We train the network on size-normalized, cutout images,
which could in a production system be provided by a person detector. Following
best practices for CNN training, we use a validation set to determine the
optimal number of training iterations and the person size, which is around
500 pixels. This high resolution allows the CNN to reliably predict small
body parts. In this challenging setup, we achieve an intersection over union
(IoU) score of 0.4432 and an accuracy of 0.9331. Qualitative results on five
datasets are shown in Fig.~\ref{fig:results_seg}.

The overall performance is compelling: even the small segments around the joints
are recovered reliably. Left and right sides of the subjects are identified
correctly, and the four parts of the head provide an estimate of head
orientation. The average IoU score is dominated by the small segments, such
as the wrists.

The VOC part dataset is a hard match for our predictor: instead of providing
instances of people, it consists of entire scenes, and many people are visible
at small scale. To provide a comparison, we use the instance annotations
from the VOC-Part dataset, cut out samples and reduce the granularity of our
segmentation to match the widely used six part representation. Because of the
low resolution of many displayed people and extreme perspectives with, \eg,
only a face visible, the predictor often only predicts the background class on
images not matching our training scheme. Still, we achieve an IoU score of 0.3185 and 0.7208 accuracy over the entire dataset without finetuning.

Additional examples from the {\LSP}, {\MPII}, {\FP}, Fashionista, VOC,
{\HumanEva} and {\HumanM} datasets are shown in the supplementary
\ifcvprfinal
material available on the project homepage\footnote{\url{http://up.is.tuebingen.mpg.de/}}.
\else
material.
\fi
The model has not been trained on any of the latter four, but
the results indicate good generalization behavior.
We include a video to visualize stability across consecutive frames.

\subsection{Human Pose Estimation\label{sec:pose}}
With the 3D body fits, we can not only generate consistent keypoints on the human skeleton but also on the body surface. For the
experiments in the rest of this paper, we designed a 91-landmark\footnote{We use
  the term `landmark' to refer to keypoints on the mesh surface to emphasize
  the difference to the so-far used term `joints' for keypoints located inside
  of the body.} set to analyze a dense keypoint set.

We distributed the landmarks according to two criteria: disambiguation of body
part configuration and estimation of body shape. The former requires
placement of markers around joints to get a good estimation of their
configuration. To satisfy the latter, we place landmarks in regular intervals
around the body to get an estimate of spatial extent independent of the
viewpoint. We visualize our selection in Fig.~\ref{fig:landmarks} and example
predictions in Fig.~\ref{fig:results_pose}.

In the visualization of predictions, we show a subset of the 91 landmarks and
only partially connect the displayed ones for better interpretability. The core 14 keypoints describing the human skeleton are part of our selection to describe the fundamental pose and maintain comparability with existing methods.

We use a state-of-the-art DeeperCut CNN~\cite{deepercut} for our pose-related
experiments, but believe that using other models such as Convolutional Pose
Machines~\cite{ConvolutionalPoseMachines:2016} or Stacked Hourglass
Networks~\cite{Newell16} would lead to similar findings.

To assess the influence of the quality of our data and the difference of the
loss function for 91 and 14 keypoints, we train multiple CNNs: (1) using all human labels
but on our (smaller) dataset for 14 keypoints ({\SMPLPose}I-P14h) and (2) on the
dense 91 landmarks from projections of the SMPL mesh ({\SMPLPose}I-P91). Again,
models are trained on size-normalized crops with cross-validated parameters. We
include the performance of the original DeeperCut CNN, which has been trained on
the full {\LSP}, {\LSPextended} and {\MPII} datasets (in total more than 52,000
people) in the comparison with the models being trained on our data (in total
5,569 people). The results are summarized in Tab.~\ref{tab:pose_results_2d}.
\begin{table}
  \begin{centering}
    \begin{tabular}{|c||c|c||c|}
      \hline
      \scriptsize{}PCK@0.2 & \scriptsize{}{\SMPLPose}I-P14h & \scriptsize{}{\SMPLPose}I-P14 & \scriptsize{}{\SMPLPose}I-P91\\
      \hline
      \hline
\scriptsize{}DeeperCut CNN~\cite{deepercut} & \scriptsize{}\textbf{93.45} & \scriptsize{}92.16 & \scriptsize{}NA \\
      \hline
      \hline
\scriptsize{}Ours (trained on {\SMPLPose}I-P14h) & \scriptsize{}89.11 & \scriptsize{}87.36 & \scriptsize{}NA \\
      \hline
      \scriptsize{}Ours (trained on {\SMPLPose}I-P91) & \scriptsize{}\textbf{91.15} & \scriptsize{}\textbf{93.24} & \scriptsize{}\textbf{93.54} \\
      \hline
    \end{tabular}
  \par\end{centering}
  \vspace*{-0.2cm}
  \caption{Pose estimation results. Even though the DeeperCut CNN has been
    trained on almost by factor ten more examples, our model remains competitive.
    The third row shows the results of our 91 landmark model evaluated on the 14
    core keypoints on human, 14 and 91 SMPL generated landmark labels. It
    outperforms the model trained on the data labeled by humans (row
    2 vs. 3, column 1) by more than two score
    points. Fair comparisons can only be made within offset boxes.\label{tab:pose_results_2d}}
\end{table}
Even though the size of the dataset is reduced by nearly an order of magnitude,
we maintain high performance compared to the original DeeperCut CNN. Comparing the two models trained on the same amount of data, we find
that the model trained on the 91 landmarks from the SMPL data has a notable
advantage of nearly six score points on the SMPL labeled data (row 2 vs. 3,
column 2). Even when evaluating on the human labeled data, it maintains an
advantage of two score points (row 2 vs. 3, column 1). This shows that the
synthetic keypoints generalize to the human labels, which we take as an
encouraging result.

We provide the third column for giving an impression of the performance of the
additional 77 landmarks. When including the additional landmarks in the
evaluation, the score rises compared to evaluating on the 14 core keypoints,
indicating their overall performance is above average. A direct comparison to the 14 keypoint values is not valid, because the score is averaged over results of differing `difficulty'.

\paragraph{Integrating a Dataset with a Different Label Set\label{sec:integrating_new_data}}
The current state-of-the art pose estimators benefit from
training on all human pose estimation datasets with a similar label set. 
The {\FP} dataset~\cite{Fashionpose:2014} would complement these well, but is annotated with a different set of keypoints: the neck joint
is missing and the top head keypoint is replaced by the nose. Due to this
difference, it is usually not included in pose estimator training.

Using our framework, we can overcome this difficulty: we adjust the
fitting objective by adding the nose to and removing the top-head keypoint from the
objective function.
We fit the SMPL model to the {\FP} dataset and curate the fits.
The additional data enlarges our training set by 1,557 images and test set by
181 images.
This forms the full {\SMPLPose}-3D
dataset, which we use for all remaining experiments.

We train an estimator on the landmarks projected from the full {\SMPLPose}-3D dataset.
This estimator outperforms the plain DeeperCut CNN with a small margin from 0.897
PCK@0.2 (DeeperCut) to 0.9028 PCK@0.2 (ours) on the full, human labeled
FashionPose test set.

\subsection{3D Human Pose Estimation\label{sec:3d_pose_estimation}}
In this section, we analyze the impact of using the 91 predicted keypoints instead of 14 for the SMPLify 3D fitting method.
For the fitting process, we rely solely on the 91 predicted 2D~landmarks
and no additional segmentation or gender information (in contrast to the SMPLify
method as described in~\cite{paper2}, where gender information is used for
fitting on the 3D datasets).
Segmentation information is not required anymore to estimate body extent due to the landmarks on the body surface.
\paragraph{{\LSP} dataset}
On the {\LSP} dataset, there is no ground truth for 3D body model fitting
available. To be independent of biases towards a specific keypoint set, we rely
on the acquired six body part segmentation to obtain meaningful performance scores
(see Tab.~\ref{tab:lsp_scores}).

The six-part manual segmentation annotations consist of head, torso, left and right
leg, and left and right arm (see Fig.~\ref{fig:seg_annotations}). While this
representation is coarse, it provides a good estimate of the overall
quality of a fit. It takes into account the body shape and not only keypoints,
hence it is a fair judge for pose estimators aiming for slightly different
keypoint locations.

\begin{table}
  \begin{center}
    \resizebox{0.48\textwidth}{!}{
    {\scriptsize{}}%
    \begin{tabular}{|c|c|c|}
      \hline
      & {\scriptsize{}FB Seg. acc., f1} & {\scriptsize{}P Seg acc., f1}\\
      \hline
      \hline
      {\scriptsize{}SMPLify on GT lms.} & {\scriptsize{}0.9176, 0.8811} & {\scriptsize{}0.8798, 0.6584}\\
      \hline
      {\scriptsize{}SMPLify on GT lms. \& GT seg.} & \textbf{\scriptsize{}0.9217, 0.8823} & \textbf{\scriptsize{}0.8882, 0.6703}\\
      \hline
      \hline
      {\scriptsize{}SMPLify on DeepCut CNN lms.~\cite{paper2}} & \textbf{\scriptsize{}0.9189, 0.8807} & \textbf{\scriptsize{}0.8771, 0.6398}\\
      \hline
      \hline
      {\scriptsize{}SMPLify on our CNN lms., tr. UPI-P14h} & {\scriptsize{}0.8944, 0.8401} & {\scriptsize{}0.8537, 0.5762}\\
      \hline
      {\scriptsize{}SMPLify on our CNN lms., tr. UP-P14} & {\scriptsize{}0.8952, 0.8475} & {\scriptsize{}0.8588, 0.5798}\\
      \hline
      {\scriptsize{}SMPLify on our CNN lms., tr. UP-P91} & \textbf{\scriptsize{}0.9099, 0.8619} & \textbf{\scriptsize{}0.8732, 0.6164}\\
      \hline
      \hline
      {\scriptsize{}DP from 14 landmarks} & {\scriptsize{}0.8649, 0.7915} & {\scriptsize{}0.8223, 0.4957}\\
      \hline
      {\scriptsize{}DP from 91 landmarks} & {\scriptsize{}0.8666, 0.7993} & {\scriptsize{}0.8232, 0.5102}\\
      \hline
      {\scriptsize{}DP from 14 lms., rotation opt.} & {\scriptsize{}0.8742, 0.8102} & {\scriptsize{}0.8329, 0.5222}\\
      \hline
      {\scriptsize{}DP from 91 lms., rotation opt.} & \textbf{\scriptsize{}0.8772, 0.8156} & \textbf{\scriptsize{}0.8351, 0.5304}\\
      \hline
    \end{tabular}{\scriptsize \par}
    }
  \end{center}

  \vspace*{-0.5cm}
  \caption{Scores of projected body parts of the fitted SMPL model on the full
    {\LSP} test set six part human labels (\textit{landmarks} is abbreviated to \textit{lms}.). Fair comparisons can only be made within offset boxes. `DP'
    refers to `Direct Prediction' (see Sec.~\ref{sec:dp}). The landmarks for these experiments are always predictions from our CNN trained on {\SMPLPose}-P91.\label{tab:lsp_scores}}
  \vspace*{-0.2cm}
\end{table}

Unsurprisingly, the segmentation scores of the SMPLify method improve when the
segmentation term (\cf~Sec.~\ref{sec:segmentation_term}) is added. Due to the
longer-trained pose estimator, SMPLify as presented in~\cite{paper2} still has
an overall advantage on the {\LSP} dataset (compare rows three and four).

Training on our generated data for 14 joints and then using SMPLify improves the scores (compare lines four and five) thanks to cleaner data and better correspondence of keypoints and SMPL skeleton. Using our 91 landmark model gives a large performance boost of 3.6 f1 score points. We do not reach the performance of the fits performed on the DeepCut CNN~\cite{Leonid2016DeepCut} predictions, largely because of few extreme poses that our pose estimator misses with a large influence on the final average score.

\vspace*{-0.1cm}
\paragraph{{\HumanEva} and {\HumanM} Datasets}
We evaluate 3D fitting on the {\HumanEva} and Human3.6M datasets where 3D ground truth keypoints are available from a motion capture system.
We follow the evaluation
protocol of SMPLify~\cite{paper2} to maintain comparability, except for subsampling to every 5th frame. This still leaves us with a
framerate of 10Hz which does not influence the scores. We do this solely due to
practical considerations, since the SMPLify fitting to 91 landmarks can take up
to twice as long as fitting to 14 keypoints. We provide a summary of results in
Tab.~\ref{tab:pose_results_3d}.

\begin{table}
  \begin{center}
    {\scriptsize{}}%
    \begin{tabular}{|c|c|c|}
      \hline
      & {\scriptsize{}{\HumanEva}} & {\scriptsize{}Human3.6M}\\
      \hline
      \hline
      {\scriptsize{}Zhou et al.~\cite{Zhou:2015b}} & {\scriptsize{}110.0} & {\scriptsize{}106.7}\\
      \hline
      {\scriptsize{}DP from 91 landmarks} & {\scriptsize{}93.5} & {\scriptsize{}93.9}\\
      \hline
      {\scriptsize{}SMPLify on DeepCut CNN lms.~\cite{paper2}} & {\scriptsize{}79.9} & {\scriptsize{}82.3}\\
      \hline
      {\scriptsize{}SMPLify on our CNN lms., tr. UPI-P14h} & {\scriptsize{}81.1} & {\scriptsize{}96.4}\\
      \hline
      {\scriptsize{}SMPLify on our CNN lms., tr. UP-P14} & {\scriptsize{}79.4} & {\scriptsize{}90.9}\\
      \hline
      {\scriptsize{}SMPLify on our CNN lms., tr. UP-P91} & \textbf{\scriptsize{}74.5} & \textbf{\scriptsize{}80.7}\\
      \hline
    \end{tabular}
  \end{center}

  \vspace*{-0.5cm}
  \caption{Average error over all joints in 3D distance
    (mm).\label{tab:pose_results_3d}}
  \vspace*{-0.2cm}
\end{table}

We do not use actor or gender specific body models, but one hybrid human model,
and rely on the additional landmarks for shape inference. This makes the
approach fully automatic and deployable to any sequence without prior knowledge.
Even with these simplifications and a magnitude fewer training examples for our
pose estimator, we achieve an improvement of 5.4mm on average on the {\HumanEva} dataset and an improvement of 1.6mm on average on the Human3.6M dataset (4th versus 5th row).

The use of a pose estimator trained on the full 91 keypoint dataset
{\SMPLPose}-P91 improves SMPLify even more. Compared to the baseline model trained
on {\SMPLPose}I-P14h, performance improves by 6.6mm on the simpler {\HumanEva}
dataset, and by 15.7mm on {\HumanM}. Even when training a 14 keypoint pose
estimator, the higher consistency of our generated labels helps to solve this
task, which becomes apparent comparing lines four and five.

\subsection{Direct 3D Pose and Shape Prediction\label{sec:dp}}
The 91~landmark predictions in 2D enable a human observer to
easily infer the 3D shape and pose of a person: the keypoints on the
body surface provide a good hint to estimate person shape and in
combination with the skeleton orientation and pose can usually be
identified (\cf Fig.~\ref{fig:results_pose}). This observation inspired us to explore the limits of 
a predictor for the 3D body model parameters directly from
the 2D keypoint input.
For this purpose, we use the 3D poses and shapes from the UP-3D dataset to
sample projected landmarks with the full 3D parameterization of SMPL as
labels. We move a virtual `camera' for every pose on 5 elevations to 36
positions around the 3D model to enhance the training
set.
On this data, we experimented with multi-layer perceptrons as well as Decision
Forests. We preferred the latter regressor, since Decision Forests are less
susceptible to noise. We train a separate forest to predict the axis-angle
rotation vector for each of the 24 SMPL joints, as well as one to predict the
depth. The input landmark positions are normalized w.r.t. position and scale to
improve generalization. We experimented with distance-based features and
dot-product features from the main skeleton connections, but these were not
as robust as plain 2D image coordinates. It turned out to be critical to use full rotation matrices as regression targets: the axis-angle representation has discontinuities, adding noise to the loss function.

One Decision Forest predicts pose or shape in ~0.13s\footnote{For all timings, a test system with a 3.2Ghz six core processor and an NVIDIA GeForce GTX970 has been used.}. The predictions of all forests
are independent, which means that the full pose and shape prediction can be obtained in between
one and two orders of magnitudes faster than with SMPLify. This could allow the
use of 3D pose and shape estimation for video applications, \eg, action recognition.

Whereas the 3D model configuration does not always match the image evidence (see
Fig.~\ref{fig:results_d3d}), it recovers the rough pose.
We provide 
scores in Tab.~\ref{tab:lsp_scores} and Tab.~\ref{tab:pose_results_3d} with the
name `DP' ({\bf D}irect {\bf P}rediction). We additionally add the scores for a
hybrid version, for which we predict pose and shape using Decision Forests
and take few optimization steps to make the global rotation of the body model
match the image evidence (with varying runtime depending on the initialization,
but less than one second on our data).

The difference between the full optimization on the {\LSP} dataset in f1 score
is 0.1062 for the 91 landmark based method and reduces with rotation
optimization to 0.086. On the 3D datasets, the direct prediction method outperforms
all optimization based methods except for SMPLify that runs in the order of
tens of seconds.

Together with our ResNet101-based CNN model, it is possible to predict a full 3D body model
configuration from an image in 0.378s. The pose-predicting CNN is the
computational bottleneck. Because our findings are not specific to a CNN
model, we believe that by using a speed-optimized CNN, such as
SqueezeNet~\cite{SqueezeNet}, and further optimizations of the direct predictor, the proposed method
could reach real-time speed.

\section{Closing the Loop\label{sec:closingtheloop}}
With the improved results for 3D fitting, which helped
to create the dataset of 3D body model fits in the first place, a natural
question is, whether the improved fitting method helps to enlarge the dataset.

\begin{figure}
  \centering
  \hspace*{0.3cm}
  \subfloat[\label{fig:f1improvement_hist}]{
    \includegraphics[width=0.4\textwidth]{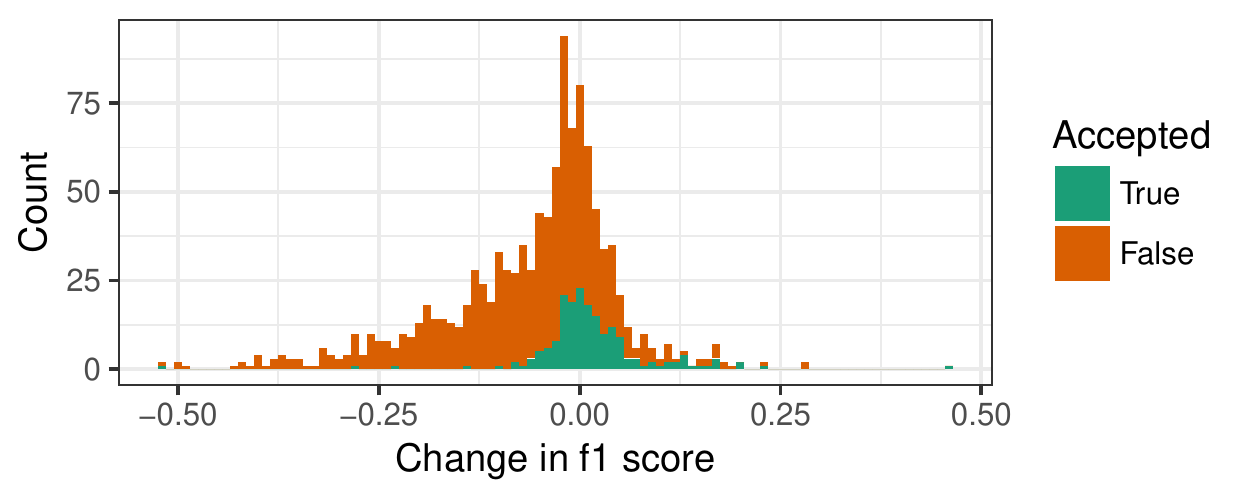}%
  }
  \\[-0.cm]
  \subfloat{
    \includegraphics[width=0.45\textwidth]{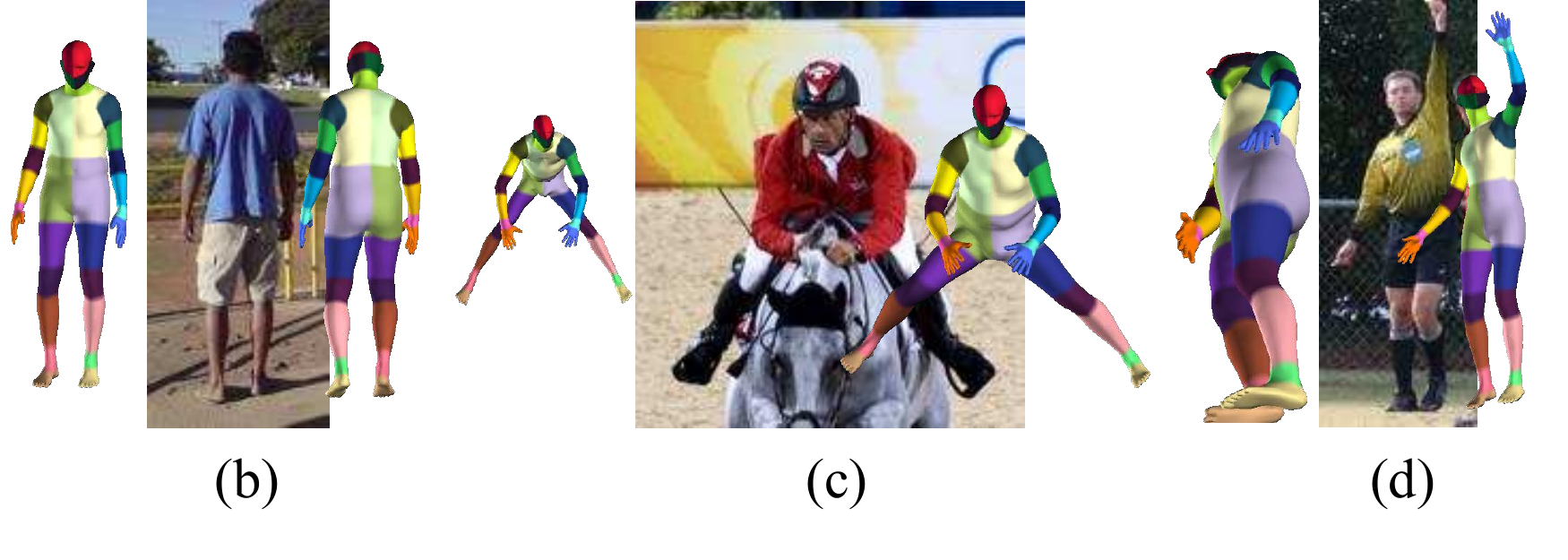}
  }
  \vspace*{-0.4cm}
  \caption{Improvement of body model fits on the unused part of the
    {\LSP} dataset. Compared are fits to 91 predicted keypoints vs. fits to 14 ground truth keypoints.
    \textbf{(a)}:~histogram of change in f1 score of projected six body part segmentation agreement
    with human annotator ground truth. Green color indicates that the formerly unaccepted fit to the 14 ground
    truth keypoints is accepted as valid when performed with the 91 predicted keypoints. For each image triple in \textbf{(b)},
    \textbf{(c)}, \textbf{(d)}: left: SMPLify fit to the 14 ground truth keypoints, right: fit to the predicted 91 landmarks from our
    predictor.\label{fig:another_iteration}}
\end{figure}

We ran SMPLify on the 91 landmark predictions from our pose estimator and again
asked human annotators to rate the 3D fits on all LSP images that were
not accepted for our initial dataset. 
Of the formerly unused 54.75\% of the data (1095 images), we found an
improvement in six body part segmentation f1 score for 308 images (\cf
Fig.~\ref{fig:f1improvement_hist}).
We show three example images with high
improvement in f1~score in Fig.~\ref{fig:another_iteration}, (b) to (d): improvement due to left-right label noise,
depth ambiguity and perspective resolution compared to fits on the 14 ground truth keypoints.
Human annotators accepted additional 185
images, which is an improvement of 20\% over the number of accepted initial fits and an
absolute improvement of 9.3\% in accepted fits of the {\LSP} dataset.

The most common reasons for improvement are (1)~noisy annotations, (2)~better
perspective resolution and (3)~the better match of keypoints to the SMPL skeleton. An even higher ratio of improvement can be expected for the datasets with more annotation noise, such as the {\LSPextended} and {\MPII} datasets. 
This enlarged set of data could be used to again train estimators and continue iteratively. 

\section{Discussion}

With the presented method and dataset, we argue for a holistic
view on human related prediction tasks. By improving the
representation of humans we could integrate datasets with different annotations
and approach established tasks at a new level of detail.
The presented results include high fidelity semantic body part
segmentation into 31 parts and 91 landmark human pose estimation. This sets a new mark in
terms of levels of detail that previous work did not reach. At the same time, it helps to
improve the state-of-the art for 3D human pose estimation on the two
standard benchmark datasets {\HumanEva} and Human3.6M.

We present a regression tree model that predicts the 3D body configuration from 2D
keypoints directly. This method runs orders of magnitude faster than
optimization based methods. This direct prediction captures the overall pose from simple 2D
input reasonably well and we are optimistic that it can be scaled to reach near
real-time performance. 
We show that the improved 3D fitting method allows more good fits that enlarge
the training set. Here, we only took one iteration but
are confident that a system that iterates over the two generative and discriminative
stages can be deployed on large scale to continuously learn and improve
with very limited human feedback.
\clearpage

{\small
\bibliographystyle{ieee}
\bibliography{bibliography}
}

\end{document}